\PassOptionsToPackage{table}{xcolor}
\documentclass[letterpaper]{article} 
\usepackage{aaai2026} 
\usepackage{times} 
\usepackage{helvet} 
\usepackage{comment}
\usepackage{courier} 
\usepackage[hyphens]{url} 
\usepackage{graphicx} 
\usepackage{natbib} 
\usepackage{caption} 
\usepackage{fontawesome5}
\newcommand{\emailmark}{%
  \textsuperscript{\raisebox{-0.15ex}{\tiny\faEnvelope}}%
}
\usepackage{booktabs}
\usepackage{multirow}
\usepackage{array}
\usepackage{amsmath}
\usepackage{amssymb}
\usepackage{tabularx}

\newcolumntype{Y}{>{\centering\arraybackslash}X}
\newcommand{\best}[1]{\textbf{#1}}
\newcommand{\second}[1]{\underline{#1}}

\title{INFANiTE: Implicit Neural Representation for High-Resolution
Fetal Brain Spatio-Temporal Atlas Learning from Clinical Thick-Slice MRI}

\author{
Xiaotian Hu\textsuperscript{\rm 1}\equalcontrib,
Mingxuan Liu\textsuperscript{\rm 1}\equalcontrib,
Hongjia Yang\textsuperscript{\rm 1}\equalcontrib,
Tongxi Song\textsuperscript{\rm 1},
Yijin Li\textsuperscript{\rm 1},
Yifei Chen\textsuperscript{\rm 1},
Haoxiang Li\textsuperscript{\rm 1},
Zihan Li\textsuperscript{\rm 1},
Yingqi Hao\textsuperscript{\rm 1},
Ziyu Li\textsuperscript{\rm 3},
Yi Liao\textsuperscript{\rm 2},
Haibo Qu\textsuperscript{\rm 2},
Qiyuan Tian\textsuperscript{\rm 1}\emailmark
}

\affiliations{
{\small
\mbox{%
\textsuperscript{\rm 1}Tsinghua University
\quad
\textsuperscript{\rm 2}Sichuan University
\quad
\textsuperscript{\rm 3}University of Oxford}
\\[-0.15em]
{\footnotesize
\faEnvelope\ \texttt{qiyuantian@tsinghua.edu.cn}}
}
}
\nocopyright
\begin{document}

\maketitle

\begin{abstract}
Spatio-temporal fetal brain atlases are important for characterizing normative neurodevelopment and identifying congenital anomalies. However, existing atlas construction pipelines necessitate days for slice-to-volume reconstruction (SVR) to generate high-resolution 3D brain volumes and several additional days for iterative volume registration, thereby rendering atlas construction from large-scale cohorts prohibitively impractical. We address these limitations with \textbf{\texttt{INFANiTE}}, an Implicit Neural Representation (INR) framework for high-resolution Fetal brain spatio-temporal Atlas learNing from clinical Thick-slicE MRI scans, bypassing both the costly SVR and the iterative non-rigid registration steps entirely, thereby substantially accelerating atlas construction. Extensive experiments show that \textbf{\texttt{INFANiTE}} achieves the best reported mean subject-consistency and reference-fidelity scores among the evaluated methods, while providing competitive intrinsic image quality and tissue-volume trajectories that broadly agree with normative developmental models. These advantages remain consistent under challenging sparse-data settings. Additionally, \textbf{\texttt{INFANiTE}} reduces the end-to-end processing time (i.e., from raw scans to the final atlas) from days to hours compared to the traditional 3D volume-based pipeline (e.g., SyGN), facilitating large-scale population-level fetal brain analysis. Code: \url{https://github.com/hu2274898/INFANiTE}

\end{abstract}

\section{Introduction}
Prenatal brain development is a critical phase with lasting implications for human neurodevelopment \cite{fetalcsr,panda}. Spatio-temporal atlases constructed from three-dimensional (3D) fetal brain MRI volumes support the identification of atypical brain patterns, providing indispensable insights into potential early manifestations of clinical conditions \cite{review}. Nevertheless, persistent fetal motion and maternal respiration preclude the direct acquisition of artifact-free 3D MRI. To address the problem, traditional fetal brain atlas construction pipelines \cite{traditional1,traditonal2,traditional4,traditional5} employ fast 2D imaging sequences (e.g., turbo spin echo, TSE) to preserve high in-plane resolution while freezing intra-shot motion. The resulting multi-planar image stacks are subsequently reconstructed into a 3D brain volume using slice-to-volume reconstruction (SVR) algorithms \cite{nesvor2023tmi,ebner2020niftymic}. Then, atlases are constructed using SyGN \cite{avants2010sygn}, an iterative framework that warps individual subjects to an evolving template through non-rigid registration, with the template being refined by averaging the warped images until convergence to achieve a stable population representation.

\begin{figure}[!t]
  \centering
  \includegraphics[width=0.88\columnwidth]{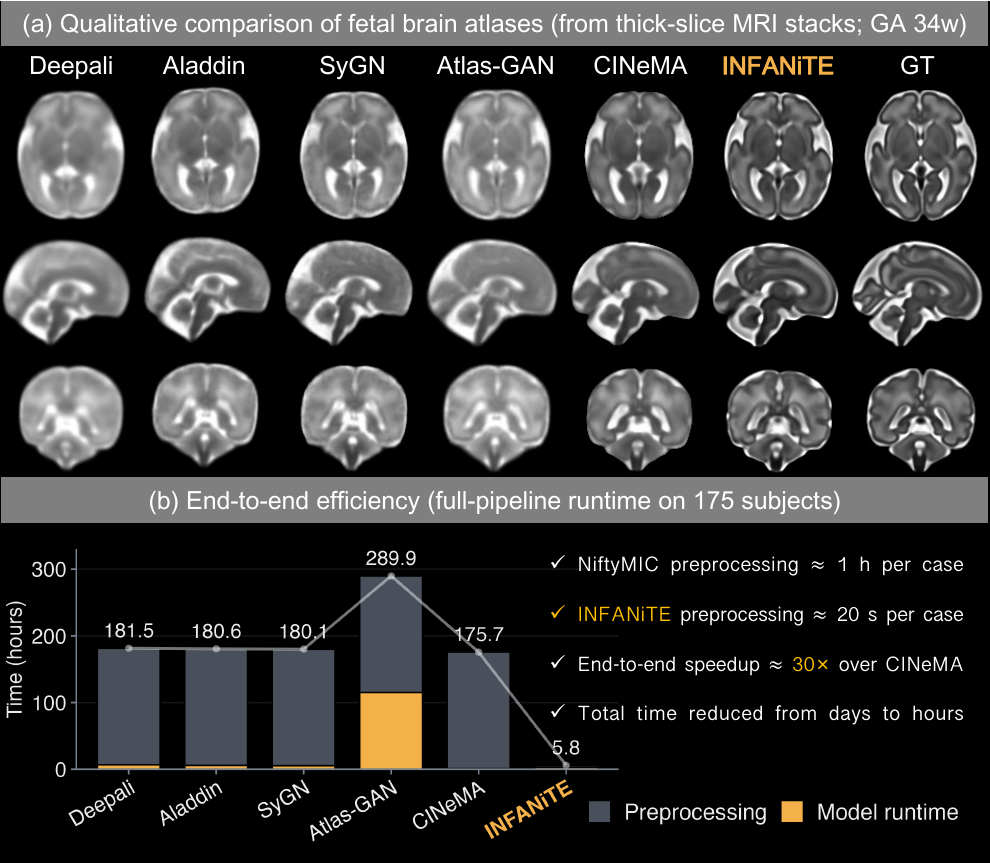}
    \caption{\textbf{(a)}~Qualitative comparison of fetal brain atlases at
  GA~34 weeks (axial, coronal, sagittal), all built directly from clinical
  thick-slice stacks (with \textbf{\texttt{INFANiTE}} preprocessing). \textbf{(b)}~End-to-end processing time on the 175-subject \textbf{Multi-Stack
  dataset}. By replacing SVR with a lightweight slice-to-template registration,
  \textbf{\texttt{INFANiTE}} cuts total time to 5.8\,h, an
  $\approx$30$\times$ speedup over CINeMA.}
  \label{fig:teaser}
\end{figure}

However, the heavy computational cost associated with both SVR and iterative registration limits the scalability of atlas construction to large cohorts representing development at the population level, thereby confining most existing anatomical fetal brain atlases to relatively small datasets of 20 to 90 volumes \cite{review}. Specifically, the widely adopted SVR algorithm NiftyMIC \cite{ebner2020niftymic} typically requires above 60 minutes to reconstruct a single fetal brain volume \cite{diagnostics13142355,10.1259/bjr.20220071}, and several recent studies have reported a reconstruction success rate ranging from only 60.0\% to 77.3\% \cite{or-kan,affirm2023tmi}. Although NeSVoR \cite{nesvor2023tmi} accelerates the SVR process, it yields lower image quality compared to NiftyMIC and proves less clinically useful than the original thick-slice scans, primarily due to increased blurring and intensity artifacts \cite{niftymic_vs_nesvor}. Moreover, its computational demand remains considerable. Following reconstruction, the iterative groupwise registration employed by SyGN \cite{avants2010sygn} further compounds the computational burden, as the number of dense registration operations scales linearly with both the cohort size and the number of iterations required for convergence, rendering large-scale atlas construction from volumetric data prohibitively expensive.

Recent deep learning-based methods bypass the iterative groupwise registration, motivated primarily by sharper templates and flexible anatomical conditioning. For instance, generative adversarial networks (GANs) recast conditional template estimation as an adversarial registration problem, yielding sharper and more condition-specific templates that better capture age- and pathology-related variability (e.g., Atlas-GAN \cite{atlas_gan} and its variants \cite{tang2025surface}). Implicit Neural Representations (INRs) instead learn continuous spatio-temporal atlases directly in latent space, dispensing with explicit deformation fields while enabling resolution-agnostic generation and conditioning on gestational age (GA) and pathologies even under limited data (e.g., CINA \cite{Dannecker2024CINACI} and CINeMA \cite{dannecker2025cinemaconditionalimplicitneural}). Notwithstanding these advances, all such methods still operate on high-resolution 3D volumes, so the time-consuming SVR preprocessing remains indispensable and the end-to-end processing time from raw clinical scans to the final atlas remains prohibitively long (175.7--289.9 hours for 175 subjects; Fig.~\ref{fig:teaser}(b)).

To maximize the acceleration of the fetal brain atlas construction pipeline, we propose \textbf{\texttt{INFANiTE}}, an implicit neural representation framework that learns high-resolution continuous spatio-temporal atlases directly from clinical thick-slice MRI stacks. Specifically, first, a robust slice-to-template registration strategy is proposed to align stacks acquired in arbitrary orientations into a standardized 3D coordinate frame. Second, physics-informed modeling of the slice acquisition process via a Point Spread Function (PSF) is incorporated alongside a spatially-weighted optimization objective, allowing the INR to learn sharp anatomical representations from sparse thick-slice observations. Third, at inference time, subject-specific latent codes are aggregated using Gaussian kernel regression, which enables the synthesis of continuous spatio-temporal atlases. Extensive experiments demonstrate that \textbf{\texttt{INFANiTE}} not only 
generates atlases with markedly sharper cortical folding and clearer 
subcortical structures than existing baselines, but also reduces the end-to-end processing time from days to hours (Fig.~\ref{fig:teaser}), enabling 
scalable population-level fetal brain analysis.

\section{Related Works}
\subsection{Traditional Atlas Construction}
Early atlas construction relied on a selected reference or a few
templates, biasing the coordinate system toward specific anatomies and limiting
population variability~\cite{avants2010sygn,Atlas3}. Groupwise methods
instead estimate a population-centered space by iteratively aligning subjects to
an evolving mean template~\cite{traditional4,avants2010sygn}, as instantiated in
early fetal atlases that registered tissue maps while modeling MR
intensity, tissue probability, and shape across GA~\cite{habas2010atlas}.
Subsequent work improved consistency through adaptive age-dependent
kernels~\cite{Atlas3} and reduced asymmetric bias via symmetric diffeomorphic
registration with GA kernel regression~\cite{traditonal2,traditional5}, and was
further extended to population- and condition-specific settings such as Chinese
fetal and spina-bifida-aperta atlases~\cite{traditional4,traditional5,fidon2024dempstershafer}.
However, these pipelines depend on SVR~\cite{ebner2020niftymic} to obtain
high-resolution 3D volumes, making construction time-consuming, memory-intensive,
and sensitive to reconstruction failures~\cite{or-kan,affirm2023tmi}, limiting
scalability to large clinical cohorts.

\subsection{Deep Learning-Based Atlas Construction}
Rather than optimizing iterative groupwise registration, deep learning methods
bypass it with neural deformation estimators~\cite{voxelmorph}, seeking sharper
templates and flexible anatomical conditioning. Along this line, hybrid schemes use
pretrained registration priors to accelerate template estimation~\cite{hybrid},
Atlas-ISTN jointly learns segmentation, registration, and a population atlas for
topological consistency~\cite{Atlas-ISTN}, and CAS-Net couples conditional atlas
generation with registration so that age-specific atlases guide fetal brain
labeling~\cite{Li2021CASNet}. GAN-based frameworks add adversarial supervision to
sharpen anatomy. Specifically, Atlas-GAN casts conditional template estimation as adversarial
registration~\cite{atlas_gan}, and later 4D infant frameworks incorporate tissue
maps to constrain intensity-only templates~\cite{atlas-ganv2,tischer2025conditional}.
INR-based methods model anatomy as a continuous coordinate-conditioned field,
with CINA encoding subject variability into latent codes conditioned on GA and
anatomy~\cite{Dannecker2024CINACI} and CINeMA extending this to multimodal,
pathology-aware perinatal atlases~\cite{dannecker2025cinemaconditionalimplicitneural}.
Nevertheless, these methods still require SVR-reconstructed 3D
volumes~\cite{ebner2020niftymic}, so their end-to-end efficiency remains bounded
by an upstream reconstruction bottleneck.

\begin{figure*}[!t]
  \centering
  \includegraphics[width=\textwidth]{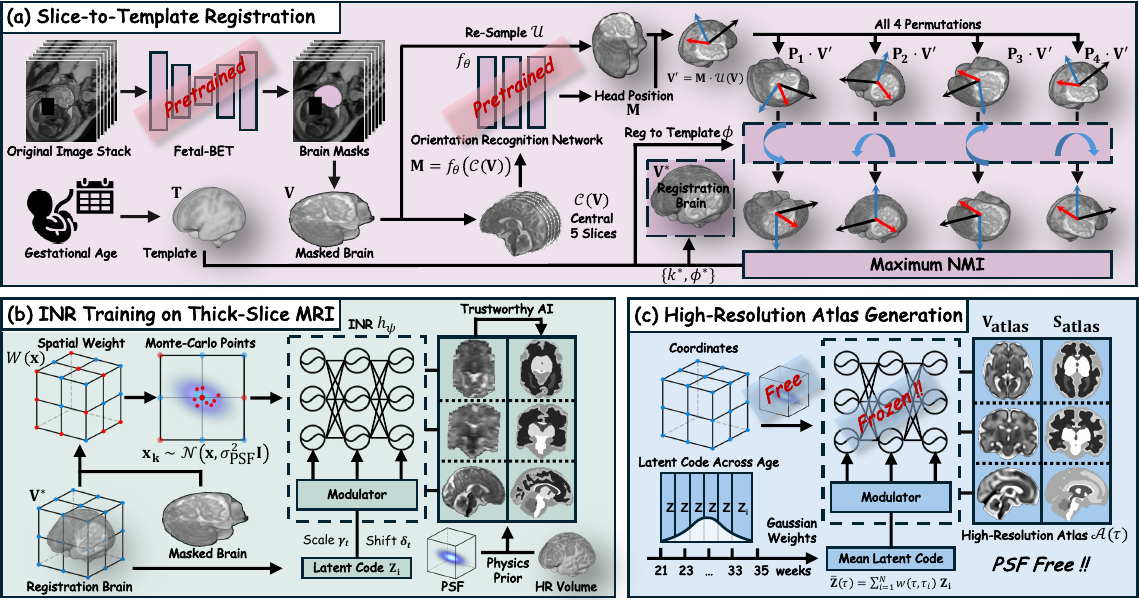}
  \caption{Overview of the proposed framework \textbf{\texttt{INFANiTE}}.
  (a) Slice-to-template registration aligns arbitrarily oriented thick-slice
  stacks to a standardized coordinate frame. (b) INR training on thick-slice MRI
  jointly optimizes a shared SIREN decoder and subject-specific latent codes
  under PSF modeling and a spatially-weighted objective. (c) High-resolution
  atlas generation queries the learned INR with age-aggregated latent codes to
  synthesize continuous spatio-temporal atlases.}
  \label{pipeline}
\end{figure*}

\section{Method}
\textbf{\texttt{INFANiTE}} reconstructs high-resolution anatomy from sparse clinical thick-slice observations, which comprises three stages (Fig. \ref{pipeline}): (a) \textbf{Slice-to-Template Registration} estimates rigid transformations $\phi^*$ to align arbitrary thick-slice stacks $\mathbf{V}_i$ into a standardized
3D coordinate frame; (b) \textbf{Implicit Representation Learning} jointly optimizes a coordinate-based network $h_\psi$ and subject-specific latent codes $\mathbf{Z}_i$ by minimizing the discrepancy between observed voxels and their physics-simulated projections; and (c) \textbf{Atlas Generation} synthesizes the atlas $\mathcal{A}(\tau)$ by querying $h_\psi$ with temporally aggregated latent features $\bar{\mathbf{Z}}(\tau)$.

\subsection{Slice-to-Template Registration}
Let $\Omega_V$ and $\Omega_T$ denote the discrete voxel grids of the brain-masked thick-slice stack $\mathbf{V}$ (using Fetal-BET \cite{fetal-bet}) and the reference template $\mathbf{T}$ \cite{traditonal2}, respectively, where $\mathbf{V}\in\mathbb{R}^{\Omega_V}$ and $\mathbf{T}\in\mathbb{R}^{\Omega_T}$.
 First, the head orientation of $\mathbf{V}$ is estimated using a pre-trained orientation recognition network \cite{or-kan} $f_\theta$, which takes the central slices $\mathcal{C}(\mathbf{V})$ as input and predicts a permutation matrix $\mathbf{M}$. 
The stack is then resampled to the isotropic resolution of $\mathbf{T}$ via operator $\mathcal{U}$, and reoriented into a canonical coordinate frame using $\mathbf{M}$, yielding the standardized stack $\mathbf{V}'$:
\begin{align}
\mathbf{V}' = \mathbf{M} \cdot \mathcal{U}(\mathbf{V}), \quad \text{where } \mathbf{M} = f_\theta(\mathcal{C}(\mathbf{V})).
\end{align}
Subsequently, combinations of $180^\circ$ rotations yield a candidate set
of four discrete orientations,
$\mathcal{P} = \{\mathbf{P}_k\}_{k=1}^4$. The optimal registered volume
$\mathbf{V}^*$ is obtained by selecting the orientation $k \in \{1,2,3,4\}$
and rigid transformation $\phi \in \Phi_{\mathrm{rigid}}$ that jointly
maximize the Normalized Mutual Information (NMI) between the template
$\mathbf{T}$ and the reoriented volume:

\begin{equation}
\begin{aligned}
\mathbf{V}^* &= \phi^*\!\left(\mathbf{P}_{k^*} \cdot \mathbf{V}'\right), \\
\{k^*, \phi^*\}
&= \operatorname*{arg\,max}_{k,\, \phi}\;
\mathrm{NMI}\!\left(
\mathbf{T},\, \phi\!\left(\mathbf{P}_k \cdot \mathbf{V}'\right)
\right).
\end{aligned}
\end{equation}

\subsection{Implicit Representation Learning}
Given a set of registered thick-slice stacks $\mathcal{D}=\{(\mathbf{V}^*_i,\mathbf{S}^*_i)\}_{i=1}^{N}$ from $N$ subjects, where $\mathbf{S}^*_i$ denotes the tissue segmentation obtained from a pre-trained trustworthy AI framework \cite{fidon2024dempstershafer}, each subject is parameterized via an INR that learns a continuous mapping from spatial coordinates $\mathbf{x}=(x,y,z)\in\Omega\subset\mathbb{R}^3$ to the corresponding intensity $\mathbf{V}^*_i(\mathbf{x}) \in \mathbb{R}$ and segmentation label $\mathbf{S}^*_i(\mathbf{x}) \in \{0,\ldots,C-1\}$ with $C$ classes.

Following SIREN \cite{sitzmann2020SIREN}, we represent the implicit signal using multilayer perceptrons (MLPs) with periodic activation functions. The weights of the MLPs are shared across the dataset to capture global anatomical structure, while a subject-specific latent code $\mathbf{Z}_i$ is optimized to modulate individual anatomical variability \cite{bauer2023spatialfunctascalingfuncta,dannecker2025cinemaconditionalimplicitneural}. Let $\mathbf{y}_t(\mathbf{x})$ denote the output feature of the $t$-th MLP layer. For any continuous coordinate $\mathbf{x}\in\Omega$, $\mathbf{z}_i(\mathbf{x})$ is queried from the latent code $\mathbf{Z}_i$ via trilinear interpolation and then used to modulate the weights of the $t$-th layer:
\begin{align}
\mathbf{z}_i(\mathbf{x}) &= \mathrm{Interp}(\mathbf{Z}_i,\mathbf{x}), \\
(\boldsymbol{\gamma}_t,\boldsymbol{\delta}_t) &= g_t\!\left(\mathbf{z}_i(\mathbf{x})\right), \\
\mathbf{y}_t(\mathbf{x}) &= \sin\!\Big(\omega_0 \cdot \boldsymbol{\gamma}_t \cdot (\mathbf{W}_t \mathbf{y}_{t-1}(\mathbf{x}) + \mathbf{b}_t) + \boldsymbol{\delta}_t\Big),
\end{align}
where $\mathbf{Z}_i \in \mathbb{R}^{D_z \times H' \times W' \times D'}$ is initialized as $\mathbf{Z}_i \sim \mathcal{N}(0,10^{-2})$ \cite{z}; $\mathrm{Interp}(\cdot)$ denotes trilinear interpolation; $g_t(\cdot)$ is a linear projection producing layer-specific scale factor $\boldsymbol{\gamma}_t$ and shift vector $\boldsymbol{\delta}_t$; $\mathbf{W}_t$ and $\mathbf{b}_t$ are shared parameters of the $t$-th MLP layer; $\mathbf{y}_0(\mathbf{x}) = \mathbf{x}$ is the input; and $\omega_0$ sets the frequency scale.

Finally, the output of the $L$-layer MLP $h_\psi(\mathbf{x}, \mathbf{z}_i(\mathbf{x})) = [\mathbf{V}_i(\mathbf{x}),\, \mathbf{S}_i(\mathbf{x})]$ is obtained by a linear projection with parameters $\mathbf{W}_{\mathrm{out}}$ and $\mathbf{b}_{\mathrm{out}}$:
\begin{equation}
\begin{aligned}
h_\psi(\mathbf{x}, \mathbf{z}_i(\mathbf{x}))
&= \mathbf{W}_{\mathrm{out}}
\big[\mathbf{y}_L(\mathbf{x});\, \mathbf{z}_i(\mathbf{x})\big]
+ \mathbf{b}_{\mathrm{out}} \\
&= \big[ \mathbf{V}_i(\mathbf{x}),\, \mathbf{S}_i(\mathbf{x}) \big],
\end{aligned}
\end{equation}
where $[\cdot\,;\,\cdot]$ denotes concatenation along the feature dimension, forming a skip connection for $\mathbf{z}_i(\mathbf{x})$.

\subsubsection{Physics-Informed PSF Modeling.}
Motivated by the partial volume effect, where limited spatial resolution causes each voxel to aggregate signals from neighboring tissues and results in image blurring \cite{rousseau2005fetalhr,kuklisova2012reconstruction,dannecker2025psf3dgs,stolt-anso2025psftrick}, a point spread function (PSF) is explicitly incorporated into the forward model to emulate physical degradation, thereby enabling the INR to learn high-resolution anatomical representations. Let $d_{i,j}(\mathbf{p})$ denote the observed intensity of pixel $\mathbf{p}$ in the $j$-th slice of stack $i$. This process is modeled as:
\begin{equation}
\begin{aligned}
d_{i,j}(\mathbf{p})
&= \int_{\mathbb{R}^3} \widetilde{\mathbf{V}}_i(\mathbf{x})\,
\kappa\!\big(\mathbf{x}-\mathcal{G}_{i,j}(\mathbf{p})\big)\, d\mathbf{x} \\
&\quad + \eta_{i,j}(\mathbf{p}),
\end{aligned}
\end{equation}
where $\widetilde{\mathbf{V}}_i(\mathbf{x})$ represents the high-resolution intensity, $\kappa(\cdot)$ is the 3D PSF, $\mathcal{G}_{i,j}(\cdot)$ maps a pixel to 3D space, and $\eta_{i,j}$ denotes noise. 

In practice, the PSF $\kappa(\cdot)$ is well approximated by a Gaussian function \cite{nesvor2023tmi}. Specifically, for a target query coordinate $\mathbf{x}\in\Omega$, a local cluster of $K$ neighboring coordinates is sampled as:
\begin{equation}
\begin{aligned}
\mathbf{x}_k &= \mathbf{x} + \boldsymbol{\epsilon}_k, \\
\boldsymbol{\epsilon}_k
&\sim \mathcal{N}\!\left(\mathbf{0},\,\sigma_{\mathrm{PSF}}^{2}\mathbf{I}\right),
\qquad k = 1, 2, \dots, K,
\end{aligned}
\end{equation}
where $\sigma_{\mathrm{PSF}}$ denotes the standard deviation of the spread.

The low-resolution output at any spatial coordinate is denoted as $\hat{h}_\psi(\mathbf{x}, \mathbf{z}_i(\mathbf{x})) = [\hat{\mathbf{V}}_i(\mathbf{x}), \hat{\mathbf{S}}_i(\mathbf{x})]$, modeled as the expectation of this continuous signal over the PSF footprint:
\begin{align}
\hat{h}_\psi(\mathbf{x}, \mathbf{z}_i(\mathbf{x}))
&=\mathbb{E}_{\mathbf{x}_k \sim \mathcal{N}(\mathbf{x},\sigma_{\mathrm{PSF}}^{2}\mathbf{I})}
\left[h_\psi\!\Big(\mathbf{x}_k,
\mathrm{Interp}(\mathbf{Z}_i,\mathbf{x}_k)\Big)\right] \nonumber\\
&\approx \frac{1}{K}\sum_{k=1}^{K} h_\psi\!\Big(
\mathbf{x}_k,\mathrm{Interp}(\mathbf{Z}_i,\mathbf{x}_k)\Big).
\end{align}

\begin{figure*}[!t]
  \centering
  \includegraphics[width=\textwidth]{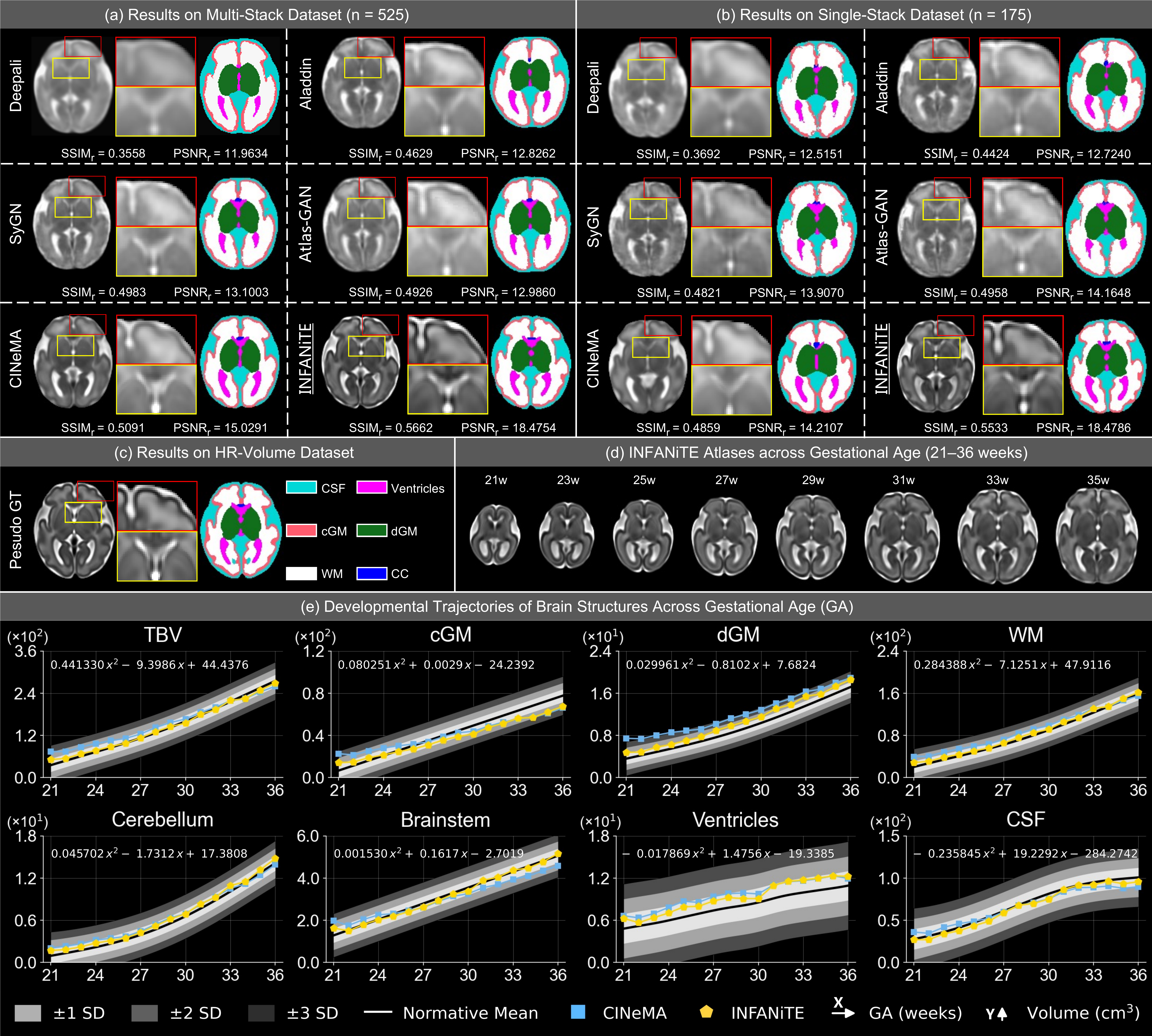}
  \caption{\textbf{(a,\,b)}~Qualitative comparison at representative GAs on
    the Multi-Stack~(\textbf{a}) and Single-Stack~(\textbf{b}) datasets. \textbf{(c)}~Pseudo GT atlas constructed from high-resolution~(HR) 3D
    volumes using CINeMA. \textbf{(d)}~\textbf{\texttt{INFANiTE}} atlas from the Multi-Stack dataset, showing smooth anatomical maturation across GA 21--36 weeks. \textbf{(e)}~Brain tissue growth trajectories for WM, cGM, dGM, cerebellum, brainstem, and TBV. Solid lines and shaded bands ($\pm$1/2/3\,SD) denote
    normative references estimated via Gaussian process regression~(GPR) fitted to
    174 training subjects. The annotated equations are second-order polynomial least-squares
    approximations of the same training data, providing closed-form summaries of
    developmental trends.}
  \label{result}
\end{figure*}

\subsubsection{Spatially-Weighted Optimization Objective.}
During registration, to distinguish truly observed voxels provided by $\mathbf{V}$ from interpolated voxels introduced by resampling, an identifier map $\mathbf{O}$ that shares the same spatial dimensions as $\mathbf{V}$ is constructed. Specifically, $\mathbf{O}$ assigns a unique index to each voxel location, establishing a one-to-one correspondence with brain voxels in $\mathbf{V}$. The identifier map $\mathbf{O}$ undergoes the same transformations as $\mathbf{V}$ throughout registration. After alignment, the transformed $\mathbf{O}$ is used to localize the observed-voxel positions in the template space, which are treated as higher-confidence supervision targets during training. A spatial weighting map $W(\mathbf{x})$ is defined such that $W(\mathbf{x})=w_{\mathrm{obs}}$ if $\mathbf{x}$ corresponds to a truly observed voxel and $W(\mathbf{x})=w_{\mathrm{int}}$ otherwise, with $w_{\mathrm{obs}} > w_{\mathrm{int}}$. The model parameters are optimized by minimizing the loss:
\begin{gather}
\mathcal{L}_{\mathrm{rec}}
= \frac{\sum_{\mathbf{x}\in\mathcal{B}} W(\mathbf{x})\,\left\lVert \hat{\mathbf{V}}_i(\mathbf{x})-\mathbf{V}_i^{*}(\mathbf{x}) \right\rVert_{1}}
{\sum_{\mathbf{x}\in\mathcal{B}} W(\mathbf{x})}, \\
\mathcal{L}_{\mathrm{seg}} = \frac{\sum_{\mathbf{x}\in\mathcal{B}} W(\mathbf{x})\, \mathcal{L}_{\mathrm{CE}}\big(\hat{\mathbf{S}}_i(\mathbf{x}), \mathbf{S}^*_i(\mathbf{x})\big)}{\sum_{\mathbf{x}\in\mathcal{B}} W(\mathbf{x})}, \\
\mathcal{L}_{\mathrm{total}} = \mathcal{L}_{\mathrm{rec}} + \lambda \mathcal{L}_{\mathrm{seg}},
\end{gather}
where $\mathcal{B}$ is a mini-batch of coordinates sampled from $\Omega$, $\mathcal{L}_{\text{CE}}$ is the cross-entropy loss, $\hat{\mathbf{S}}_i(\mathbf{x})\in\mathbb{R}^C$ denotes the predicted tissue probability, and $\lambda=1$.

\subsection{High-Resolution Atlas Generation}

During the inference phase, the PSF simulation is disabled, allowing us to directly query the learned INR for high-resolution anatomy. For spatio-temporal atlas generation at any GA $\tau$, the network is modulated by an age-specific latent code $\bar{\mathbf{Z}}(\tau)$, computed via Gaussian kernel regression over the subject codes $\{\mathbf{Z}_i\}_{i=1}^N$ with a temporal bandwidth $\sigma_\tau$:
\begin{equation}
\begin{aligned}
\bar{\mathbf{Z}}(\tau)
&= \sum_{i=1}^{N} w(\tau,\tau_i)\,\mathbf{Z}_i \\
&= \frac{\sum_{i=1}^{N} k(\tau, \tau_i)\,\mathbf{Z}_i}
{\sum_{i=1}^{N} k(\tau, \tau_i)}, \\
k(\tau, \tau_i) &= e^{-\frac{(\tau - \tau_i)^2}{2\sigma_\tau^2}},
\end{aligned}
\end{equation}
where $\tau_i$ denotes the GA of subject $i$, and $w(\tau, \tau_i)$ denotes the normalized kernel weight. 
The atlas $\mathcal{A}(\tau)$ at GA $\tau$ is obtained by querying the INR at each spatial coordinate $\mathbf{x}$ with the interpolated latent code $\bar{\mathbf{z}}(\tau,\mathbf{x})=\mathrm{Interp}(\bar{\mathbf{Z}}(\tau),\mathbf{x})$:
\begin{equation}
\mathcal{A}(\tau, \mathbf{x}) = \big[\mathbf{V}_{\text{atlas}}(\mathbf{x}),\, \mathbf{S}_{\text{atlas}}(\mathbf{x})\big]
=
h_\psi\!\big(\mathbf{x},\, \bar{\mathbf{z}}(\tau,\mathbf{x})\big).
\end{equation}

\newcommand{\std}[1]{\ensuremath{_{\pm #1}}}
\renewcommand{\best}[1]{\cellcolor{blue!18}#1}
\renewcommand{\second}[1]{\cellcolor{black!12}#1}

\begin{table*}[!t]
\centering
\resizebox{\textwidth}{!}{%
\begin{tabular}{@{} l *{8}{c} @{} }
\toprule
\textbf{Method} & \textbf{HD$_{95}$~$\downarrow$} & \textbf{ASD} & \textbf{DSC} & \textbf{TCT}
& \textbf{PSNR$_\text{r}$} & \textbf{SSIM$_\text{r}$}
& \textbf{PSNR$_\text{a}$} & \textbf{SSIM$_\text{a}$} \\
\midrule
\multicolumn{9}{l}{\cellcolor{gray!10}\textit{(a) Multi-Stack Dataset}} \\
Deepali     & 3.2413\std{0.4028} & 1.1774\std{0.1071} & 0.7149\std{0.0332} & 0.2627\std{0.0837} & 11.9634\std{1.1059} & 0.3558\std{0.0360} & 13.5951\std{0.7007} & 0.4091\std{0.0746} \\
Atlas-GAN   & 2.9080\std{0.3086} & 1.0854\std{0.0713} & 0.7348\std{0.0310} & 0.1590\std{0.0979} & 12.9860\std{1.5504} & 0.4926\std{0.0310} & 16.1068\std{1.1377} & 0.6228\std{0.0509} \\
SyGN        & 2.8669\std{0.3081} & 1.0661\std{0.0733} & 0.7395\std{0.0300} & 0.1330\std{0.0867} & 13.1003\std{1.1712} & 0.4983\std{0.0322} & 16.2402\std{0.8075} & 0.6138\std{0.0703} \\
Aladdin     & 2.6061\std{0.3714} & 0.9322\std{0.1083} & 0.7711\std{0.0333} & 0.5650\std{0.2284} & 12.8262\std{0.5475} & 0.4629\std{0.0286} & 15.6580\std{1.0584} & 0.5674\std{0.0971} \\
CINeMA      & \second{2.4137}\std{0.2545} & \second{0.8766}\std{0.0705} & \second{0.7849}\std{0.0306} & \best{1.0133}\std{0.1718} & \second{15.0291}\std{0.9650} & \second{0.5091}\std{0.0285} & \second{19.5325}\std{0.3888} & \second{0.7073}\std{0.0363} \\
\textbf{\texttt{INFANiTE}} & \best{2.3572}\std{0.2402} & \best{0.8552}\std{0.0736} & \best{0.7865}\std{0.0331} & \second{0.9702}\std{0.2832} & \best{18.4754}\std{0.6242} & \best{0.5662}\std{0.0358} & \best{22.3165}\std{1.4984} & \best{0.7854}\std{0.0403} \\
\addlinespace[4pt]
\multicolumn{9}{l}{\cellcolor{gray!10}\textit{(b) Single-Stack Dataset}} \\
Deepali     & 3.2914\std{0.5354} & 1.2065\std{0.1605} & 0.7013\std{0.0452} & 0.2794\std{0.0938} & 12.5151\std{1.5311} & 0.3692\std{0.0381} & 14.1211\std{0.8682} & 0.4239\std{0.0720} \\
Atlas-GAN   & 3.0763\std{0.5130} & 1.1366\std{0.1419} & 0.7191\std{0.0457} & 0.1638\std{0.0969} & 14.1648\std{1.8097} & 0.4958\std{0.0231} & 17.1474\std{1.1221} & 0.6233\std{0.0710} \\
SyGN        & 3.0873\std{0.5104} & 1.1406\std{0.1639} & 0.7175\std{0.0520} & 0.1556\std{0.1069} & 13.9070\std{1.6264} & 0.4821\std{0.0293} & 16.7512\std{0.8505} & 0.5869\std{0.0840} \\
Aladdin     & \second{2.7356}\std{0.2750} & \second{0.9733}\std{0.0810} & 0.7536\std{0.0281} & 0.5226\std{0.1303} & 12.7240\std{0.5539} & 0.4424\std{0.0234} & 15.2615\std{0.9458} & 0.5307\std{0.0861} \\
CINeMA      & 2.7548\std{0.3602} & 0.9860\std{0.1181} & \second{0.7663}\std{0.0468} & \best{0.9308}\std{0.2334} & \second{14.2107}\std{0.9323} & \second{0.4859}\std{0.0213} & \second{18.0846}\std{0.7249} & \second{0.6507}\std{0.0491} \\
\textbf{\texttt{INFANiTE}} & \best{2.5394}\std{0.2829} & \best{0.9153}\std{0.0874} & \best{0.7721}\std{0.0394} & \second{0.8143}\std{0.2368} & \best{18.4786}\std{0.6871} & \best{0.5533}\std{0.0378} & \best{21.6521}\std{1.3835} & \best{0.7593}\std{0.0424} \\
\bottomrule
\end{tabular}%
}
\caption{Image quality on the Multi-Stack and Single-Stack datasets, reported as mean\std{std}. Subject consistency is measured by HD$_{95}$ (mm), ASD (mm), DSC and PSNR/SSIM against test subjects (subscript r); intrinsic quality by TCT; and reference fidelity by PSNR/SSIM against the pseudo ground-truth atlas (subscript a). Best in \colorbox{blue!18}{blue}, runner-up in \colorbox{black!12}{gray}.}
\label{tab:image_quality}
\end{table*}

\begin{table*}[!t]
\centering
\resizebox{\textwidth}{!}{%
\begin{tabular}{@{} l *{8}{c} @{} }
\toprule
\textbf{Method} & \textbf{TBV} & \textbf{WM} & \textbf{BS} & \textbf{cGM} & \textbf{dGM} & \textbf{CSF} & \textbf{Cereb.} & \textbf{Vent.} \\
\midrule
Deepali     & 17.1869\std{6.6771} & 17.0883\std{7.0347} & \second{0.1136}\std{0.1294} & \second{4.1339}\std{3.1167} & 2.9608\std{0.8561} & 6.9663\std{3.9245} & \best{0.2937}\std{0.2499} & 0.7056\std{0.3550} \\
Atlas-GAN   & 18.2886\std{5.7141} & 16.8883\std{5.6128} & 0.1418\std{0.1181} & \best{3.4295}\std{2.2168} & 2.4323\std{0.4909} & 7.4959\std{3.9253} & 0.3503\std{0.3041} & \second{0.6559}\std{0.3683} \\
SyGN        & 10.4883\std{4.1709} & 12.7053\std{4.5713} & 0.2587\std{0.1153} & 4.7844\std{3.9497} & 1.8042\std{0.4311} & \best{2.8199}\std{2.0236} & \second{0.3089}\std{0.2963} & \best{0.6249}\std{0.4190} \\
Aladdin     & \best{4.2072}\std{3.8226} & \second{4.2575}\std{2.8899} & 0.3199\std{0.0982} & 5.9413\std{4.0827} & \second{1.7586}\std{0.8774} & 6.8454\std{4.8887} & 0.3943\std{0.4620} & 1.1957\std{0.6668} \\
CINeMA      & 15.3309\std{9.5437} & 9.3551\std{5.0896} & 0.2904\std{0.1718} & 6.9991\std{4.4197} & 2.6446\std{0.5225} & 4.8112\std{3.3774} & 0.6120\std{0.2866} & 1.6865\std{0.3318} \\
\textbf{\texttt{INFANiTE}} & \second{4.9467}\std{3.9090} & \best{3.7115}\std{2.1953} & \best{0.0941}\std{0.0845} & 5.8308\std{3.3250} & \best{1.1852}\std{0.3249} & \second{3.0101}\std{2.1943} & 0.4880\std{0.2013} & 1.3118\std{0.4535} \\
\bottomrule
\end{tabular}%
}
\caption{Biological plausibility, reported as mean\std{std} L1 error (cm$^{3}$, $\downarrow$) between atlas-derived tissue volumes and normative developmental trajectories. TBV: Total Brain Volume; WM: White Matter; BS: Brainstem; cGM: cortical Gray Matter; dGM: deep Gray Matter; CSF: Cerebrospinal Fluid; Cereb.: Cerebellum; Vent.: Ventricles. Best in \colorbox{blue!18}{blue}, runner-up in \colorbox{black!12}{gray}.}
\label{tab:bio_plausibility}
\end{table*}

\section{Experiments}
\subsection{Evaluation Settings}
\subsubsection{Data Acquisition.}
A total of 615 clinical 2D T$_2$-weighted thick-slice MRI scans were acquired in orthogonal planes (i.e., axial, coronal, sagittal) from 205 pregnant women with normal fetuses (GA: 21--36 weeks) on a 1.5T Philips Achieva scanner with a 16-channel body coil using a TSE sequence. The acquisition parameters spanned an echo time of 70--120ms, a repetition time of 467--15{,}000ms, a flip angle of 75--90$^\circ$, an in-plane resolution ranging from 0.39$\times$0.39 to 0.98$\times$0.98mm$^2$, 10--52 slices per stack, a slice thickness of 3--5mm, and a slice spacing of 3--6.1mm. For atlas construction, 525 stacks from 175 subjects were used as the \textbf{Multi-Stack Dataset}. Additionally, to evaluate the robustness of each method under a more challenging and clinically realistic sparse-data scenario where only a single scan per subject is available, a \textbf{Single-Stack Dataset} was curated by randomly selecting one scan per subject from the \textbf{Multi-Stack Dataset}. The remaining 90 stacks from 30 subjects served as the test dataset, for which high-resolution 3D volumes were generated using NiftyMIC \cite{ebner2020niftymic} and segmented via a pre-trained trustworthy AI framework \cite{fidon2024dempstershafer} to evaluate image quality.

\subsubsection{Implementation Details.}
\textbf{\texttt{INFANiTE}} was implemented in PyTorch and trained on a single NVIDIA H100 GPU with 80~GB memory, hosted on a server with dual Intel Xeon Platinum 8362 CPUs (2.80~GHz, 64 cores in total). The model was trained for 5 epochs with a subject batch size of 250. In each training iteration, 18,000 3D coordinates were randomly sampled, and 16 dataloader workers were used for data loading.
The implicit decoder was implemented as a SIREN-based multilayer perceptron (MLP) with a hidden dimension of 1024 and 5 hidden layers. The decoder was modulated by a subject-specific latent code with size $256 \times 3 \times 3 \times 3$ at layers 1, 3, and 5, with the SIREN frequency parameters set to $\omega=(30,30)$.
To model acquisition blur, we applied PSF-based Monte Carlo averaging by querying the decoder at Gaussian-perturbed coordinates and averaging the corresponding predictions. The number of Monte Carlo samples was scheduled as $P=4$ for epoch 0, $P=8$ for epoch 1, and $P=16$ for epochs $\geq 2$. The Gaussian perturbation standard deviation was set to $\sigma=(0.02,0.02,0.02)$.
During inference, isotropic atlases were generated at 0.8~mm spacing across 21--36 weeks GA.

\begin{table*}[!t]
\centering
\resizebox{\textwidth}{!}{%
\begin{tabular}{@{} l *{8}{c} @{} }
\toprule
\textbf{Method} & \textbf{HD$_{95}$} & \textbf{ASD} & \textbf{DSC} & \textbf{TCT}
& \textbf{PSNR$_\text{r}$} & \textbf{SSIM$_\text{r}$}
& \textbf{PSNR$_\text{a}$} & \textbf{SSIM$_\text{a}$} \\
\midrule
w/o PSF        & 2.4366\std{0.2853} & 0.8762\std{0.1043} & \best{0.7884}\std{0.0418} & \best{1.2014}\std{0.2661} & 17.1436\std{0.6367} & 0.4976\std{0.0294} & 20.3401\std{1.0307} & 0.6802\std{0.0562} \\
w/o Weight     & 2.4540\std{0.1952} & 0.8775\std{0.0484} & 0.7768\std{0.0240} & 0.8280\std{0.1962} & \best{18.6691}\std{0.7484} & \second{0.5620}\std{0.0355} & \second{22.0845}\std{1.3671} & \second{0.7802}\std{0.0361} \\
Random Weight  & \second{2.4273}\std{0.2807} & \second{0.8672}\std{0.0819} & 0.7819\std{0.0370} & 0.8527\std{0.2309} & \second{18.6488}\std{0.6390} & 0.5562\std{0.0438} & 21.9457\std{1.4822} & 0.7740\std{0.0476} \\
\textbf{\texttt{INFANiTE}} & \best{2.3572}\std{0.2402} & \best{0.8552}\std{0.0736} & \second{0.7865}\std{0.0331} & \second{0.9702}\std{0.2832} & 18.4754\std{0.6242} & \best{0.5662}\std{0.0358} & \best{22.3165}\std{1.4984} & \best{0.7854}\std{0.0403} \\
\bottomrule
\end{tabular}%
}
\caption{Ablation study on the Multi-Stack Dataset, reported as mean\std{std}, isolating the contribution of PSF modeling and the spatially-weighted objective. Metrics follow Table~\ref{tab:image_quality}. Best in \colorbox{blue!18}{blue}, runner-up in \colorbox{black!12}{gray}.}
\label{tab:ablation}
\end{table*}

\subsubsection{Baseline Methods.}
For fairness, all baselines receive the same thick-slice stacks aligned by our
slice-to-template registration (Section 2.1), so differences reflect atlas
modeling rather than preprocessing. We compare against:
(i) optimization-based SyGN \cite{avants2010sygn} and Deepali \cite{schuh2023deepali};
(ii) GAN-based Atlas-GAN \cite{atlas_gan}; and (iii) learning-based Aladdin \cite{ding2022aladdin}
and CINeMA \cite{dannecker2025cinemaconditionalimplicitneural}, the latter being our
closest competitor.

\subsubsection{Evaluation Metrics.} (1) \underline{Subject Consistency} assesses how faithfully the atlas represents the population. The generated atlas and its segmentation are registered to each test subject, yielding the Peak Signal-to-Noise Ratio (PSNR$_\mathrm{r}$) and Structural Similarity Index Measure (SSIM$_\mathrm{r}$), together with segmentation metrics comprising the Dice Coefficient (DSC), Hausdorff Distance (HD$_{95}$), and Average Surface Distance (ASD). (2) \underline{Reference Fidelity} quantifies the deviation from the pseudo ground truth (GT). As CINeMA constitutes the state-of-the-art atlas construction method built on high-resolution 3D volumes, we derive the pseudo GT reference atlas by training CINeMA on SVR-reconstructed brain volumes from the \textbf{Multi-Stack Dataset}, and report PSNR$_\mathrm{a}$ and SSIM$_\mathrm{a}$ against this reference. We emphasize that the reference atlas and the CINeMA baseline are two distinct model instances. The reference is CINeMA trained on NiftyMIC-reconstructed high-resolution volumes, whereas the CINeMA baseline in Table 1 is trained on the same thick-slice stacks as all other methods. The former serves as a proxy for the full high-cost pipeline, not as anatomical GT, since artifact-free 3D fetal MRI is unobtainable. (3) \underline{Intrinsic Quality} is measured via TCT \cite{sun2025foundation_mri} to characterize the contrast between white matter (WM) and gray matter (GM). (4) \underline{Biological Plausibility} on the \textbf{Multi-Stack Dataset} evaluates conformity to typical growth patterns through the L1 error (cm$^3$) between atlas-derived tissue volumes and normative developmental trajectories.

\subsection{Results}

\subsubsection{Qualitative Evaluation.}
\textbf{\texttt{INFANiTE}} produced sharper atlases with clearer cortical boundaries and more distinct internal structures, whereas competing methods exhibited varying degrees of blurring (Fig.~\ref{result}a, b). Compared with the pseudo GT atlas (Fig.~\ref{result}c), \textbf{\texttt{INFANiTE}} demonstrated higher visual similarity in both global appearance and local structural patterns. These visual gains remained consistent across the axial, sagittal, and coronal views, reflecting the isotropic anatomical fidelity of \textbf{\texttt{INFANiTE}}. Importantly, the \textbf{\texttt{INFANiTE}} atlas exhibited smooth and continuous anatomical maturation trajectories that align significantly closer to normative references than CINeMA (Fig.~\ref{result}d, e).

\begin{figure}[!h]
  \centering
  \includegraphics[width=0.9\columnwidth]{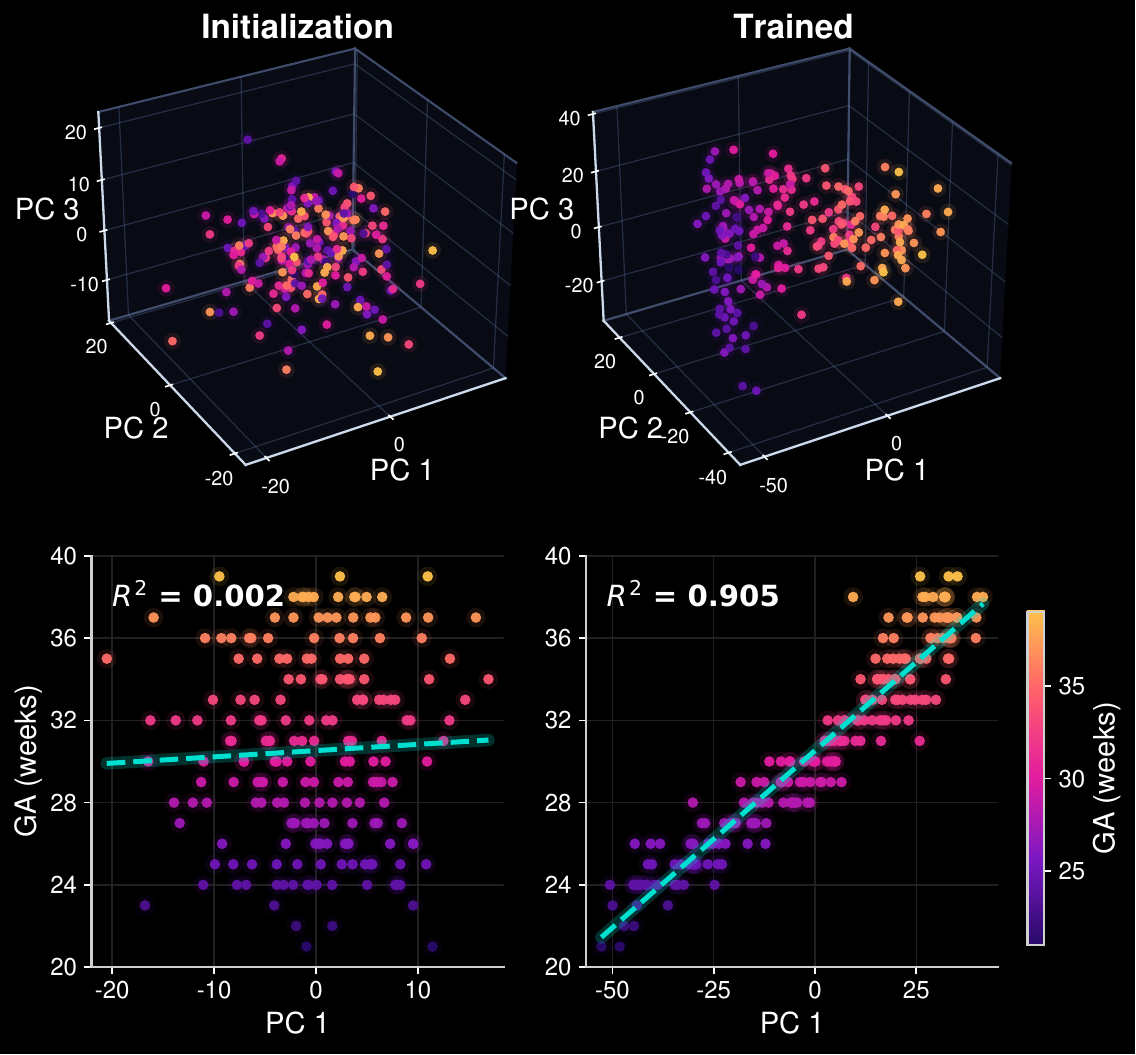}
  \caption{PCA of the subject-specific latent codes $\{\mathbf{Z}_i\}$ before and after training on the
  Single-Stack dataset.} 
  \label{fig:tsne}
\end{figure}

\subsubsection{Image Quality and Fidelity.}
Quantitatively, \textbf{\texttt{INFANiTE}} achieved the best overall image
quality on both datasets (Table~\ref{tab:image_quality}). On the Multi-Stack
Dataset, it reached HD$_{95}$ 2.3572\,mm, ASD 0.8552\,mm, and DSC 0.7865,
improving over the runner-up CINeMA (HD$_{95}$ 2.4137, ASD 0.8766, DSC 0.7849)
by 2.3\%, 2.4\%, and 0.2\%, respectively. The gains were substantially larger
on the intensity-based subject-consistency metrics, where
\textbf{\texttt{INFANiTE}} obtained PSNR$_\mathrm{r}$ 18.4754 and
SSIM$_\mathrm{r}$ 0.5662, exceeding CINeMA (15.0291 / 0.5091) by 22.9\% and
11.2\%. For reference fidelity against the CINeMA-derived pseudo GT,
\textbf{\texttt{INFANiTE}} yielded PSNR$_\mathrm{a}$ 22.3165 and
SSIM$_\mathrm{a}$ 0.7854, corresponding to gains of 14.3\% and 11.0\% over
CINeMA (19.5325 / 0.7073), indicating that atlases learned directly from
thick-slice stacks approach those built on costly SVR volumes. Under the more
challenging Single-Stack setting, \textbf{\texttt{INFANiTE}} preserved its
advantage (HD$_{95}$ 2.5394, DSC 0.7721, PSNR$_\mathrm{a}$ 21.6521,
SSIM$_\mathrm{a}$ 0.7593), with only marginal degradation relative to the
Multi-Stack results, confirming its robustness when only a single scan per
subject is available. On the intrinsic-quality metric TCT,
\textbf{\texttt{INFANiTE}} ranked second (0.9702 on Multi-Stack, 0.8143 on
Single-Stack), closely following CINeMA (1.0133 / 0.9308). We attribute
CINeMA's marginally higher WM--GM contrast to its tendency toward
over-smoothed, high-contrast templates, which is not necessarily desirable
given its weaker performance on fidelity and consistency metrics.

\subsubsection{Biological Plausibility.}
The tissue-volume analysis (Table~\ref{tab:bio_plausibility}; Fig.~\ref{result}e)
shows that \textbf{\texttt{INFANiTE}} best conforms to normative developmental
trajectories. It achieved the lowest L1 errors in WM (3.7115\,cm$^3$), BS
(0.0941\,cm$^3$), and dGM (1.1852\,cm$^3$), and ranked second in TBV
(4.9467\,cm$^3$) and CSF (3.0101\,cm$^3$), closely trailing the respective
best methods. These results indicate that the atlases produced
by \textbf{\texttt{INFANiTE}} not only look sharper but also encode
biologically faithful volumetric growth, a prerequisite for downstream
normative modeling and anomaly detection.

\subsubsection{Latent Space Analysis.}
We visualize the optimized codes $\{\mathbf{Z}_i\}$ of the Single-Stack dataset
via PCA before and after training (Fig.~\ref{fig:tsne}). Initially drawn from
$\mathcal{N}(0,10^{-2})$, the codes form a GA-agnostic cloud with PC~1 carrying
no age information ($R^2=0.002$). After joint optimization, they self-organize
into a smooth, GA-ordered manifold with PC~1 strongly correlated with GA
($R^2=0.905$). This structure emerges without any age-regression loss, purely
from the shared SIREN decoder, justifying the Gaussian kernel regression over
$\{\mathbf{Z}_i\}$ for synthesizing age-conditioned atlases.

\subsubsection{Ablation Study.}
We isolated the two key components of \textbf{\texttt{INFANiTE}} on the
Multi-Stack Dataset (Table~\ref{tab:ablation}). Removing PSF modeling caused
the sharpest drop in fidelity, lowering PSNR$_\mathrm{a}$ from 22.3165 to
20.3401 and SSIM$_\mathrm{a}$ from 0.7854 to 0.6802, which confirms that
explicitly emulating acquisition blur is essential for recovering
high-resolution anatomy from thick slices. Removing the spatially-weighted
objective reduced DSC from 0.7865 to 0.7768 and SSIM$_\mathrm{a}$ from 0.7854
to 0.7802, indicating that up-weighting truly observed voxels provides more
reliable supervision than treating all resampled voxels equally. Replacing the
observation-guided weights with random weighting produced intermediate results
(DSC 0.7819, SSIM$_\mathrm{a}$ 0.7740), further isolating the benefit of the
identifier-map--driven weighting from that of non-uniform weighting alone.

\section{Conclusion}
We presented \textbf{\texttt{INFANiTE}}, an implicit neural representation framework that constructs high-resolution fetal brain atlases from clinical thick-slice MRI, bypassing both SVR and iterative groupwise registration, reducing end-to-end processing time from days to hours, and establishing a scalable pathway toward population-level fetal brain analysis from routinely acquired thick-slice scans in real-world clinical practice.
\bibliography{mybibliography}
\end{document}